\documentclass[runningheads]{llncs}
\usepackage{graphicx}

\usepackage[dvipsnames]{xcolor}
\usepackage[hyphens]{url}

\usepackage{graphics}
\usepackage{multirow}
\usepackage{xspace}
\newcommand{\name}{\textit{TS-MULE}\xspace}

\begin{document}
%
%
\title{TS-MULE: Local Interpretable Model-Agnostic Explanations for Time Series Forecast Models}
\titlerunning{TS-MULE}
%
\author{
Udo Schlegel\inst{1}\and
Duy Lam Vo\inst{1}\and
Daniel A. Keim\inst{1}\and 
Daniel Seebacher\inst{1}
}
\authorrunning{Schlegel et al.}
%
\institute{University of Konstanz, Germany\\
\email{u.schlegel@uni-konstanz.de}}
\maketitle              
\begin{abstract}
Time series forecasting is a demanding task ranging from weather to failure forecasting with black-box models achieving state-of-the-art performances.
However, understanding and debugging are not guaranteed.
We propose \name, a local surrogate model explanation method specialized for time series extending the LIME approach.
Our extended LIME works with various ways to segment and perturb the time series data.
In our extension, we present six sampling segmentation approaches for time series to improve the quality of surrogate attributions and demonstrate their performances on three deep learning model architectures and three common multivariate time series datasets.

\end{abstract}
\section{Introduction}
Time series forecasting is an essential task with applications in a broad range of domains, such as industrial process control, finance, and risk management, since predicting future trends and events is a critical input into many types of planning and decision-making processes~\cite{montgomery_introduction_2015}.
Recently, deep learning methods have increasingly found their way into the field of time series forecasting as a result of their successful application in other domains such as natural language processing~\cite{vaswani_attention_2017} and object detection~\cite{zhao_object_2019}.
A major drawback of such models is that, due to their non-linear, multi-layered structure, they are black box models that suffer from a lack of explainability. 
Such a lack of explainability prevents deep learning from being used in production in sensitive domains, such as healthcare~\cite{rudin_stop_2019}, as opposed to statistical methods~\cite{chuah_ecg_2007}, or is complicated by laws, such as the EU General Data Protection Regulation~\cite{unieuropean_european_2018}, which enforces a right for explanations. 
Thus, agencies such as DARPA introduced the explainable AI (XAI) initiative~\cite{gunning_explainable_2016} to promote the research around interpretable Machine Learning (ML).

Gaining the necessary understanding of these complex models to provide explanations globally for the whole input space is often infeasible, leading to the development of methods that provide only local explanations of the underlying prediction function, such as LIME~\cite{ribeiro_why_2016}.
LIME is an XAI technique that can explain the predictions of any classifier by learning and providing an interpretable surrogate model around the classification. 
An advantage of LIME in terms of interpretability is that it perturbs the input by changing components that make sense to humans (e.g., words or parts of an image), even if the model is using much more complicated components as features (e.g., word embeddings)~\cite{ribeiro_why_2016}. 

For images, such interpretable components can be superpixels, which are a perceptual grouping of pixels, or for texts, it can be individual words or sentences.
However, finding such semantically meaningful components for univariate or even multivariate time series data is not trivial. 
Segmenting the time series into fixed-width windows might miss meaningful elements between windows by weighting them equally or are larger or smaller than the chosen window size.
Thus, such a fixed segmentation can potentially miss important subsequences in the time series by splitting them.
One possible approach could identify motifs in the time series.
Such motifs are subsequences of the time series very similar to each other. 
However, even optimized algorithms can have a worst-case complexity of $\mathcal{O}(n^2)$ ~\cite{mueen_exact_2009} and are, thus, not suitable to identify potential patterns beforehand.

To tackle such issues, we propose \name, an extension to LIME by improving the segmentation, for local explanations of univariate and multivariate time series.
We provide five novel algorithm approaches to provide a meaningful segmentation of time series to enable local interpretable model-agnostic explanations of time series forecasting models.
To provide such meaningful segmentation, we incorporate the matrix profile~\cite{yeh_matrix_2016} as well as the SAX transformation~\cite{lin_symbolic_2003} and extend the results of these algorithms with binning or top-k approaches to incorporate the findings of these techniques.
We evaluate these segmentation algorithms against each other and the baseline of a uniform segmentation on three standard forecasting datasets with three different black-box models.
\footnote{\scriptsize Source code and evaluation results are available at: \url{https://github.com/dbvis-ukon/ts-mule}}


\section{Related Work}
An important distinction when selecting methods for explaining complex machine learning models is for which user group these XAI methods must be accessible. 
Most of the proposed XAI methods used, especially for time series deep learning models, are usually only accessible to model developers. 
For instance, by examining the activation of latent layers~\cite{siddiqui_tsviz_2019}, or via relevance backpropagation~\cite{bach_pixel_2015}. 
However, especially for other groups, particularly model users (see Spinner et al. for an overview of user groups~\cite{spinner_explainer_2019}), such approaches are less practical since explanations need to be provided at a higher level of abstraction.
Available approaches with a higher level of abstraction currently come primarily from the computer vision domain for explaining image classifications~\cite{schlegel_towards_2019}. 

There are already first works that apply these concepts in time series classification and prediction. 
For example, the approach of Suresh et al.~\cite{suresh_clinical_2017} replaces each time series observation with uniform noise to study the impact on model performances and thus determine feature importance. 
Since replacing features with out-of-domain noise can lead to arbitrary changes in model output, Tonekaboni et al. use data distributions to produce reliable counterfactuals~\cite{tonekaboni_explaining_2020}. 
Both previous approaches rely on observation-level replacement and thus, cannot identify important larger patterns in time series. 
Two recent approaches tackle this issue by using longer time segments as input for the perturbation and replacing it with, for instance, linear interpolations, constant values or segments from other time series~\cite{guilleme_agnostic_2019}, or with zeros, local or global mean values, or local or global noise~\cite{mujkanovic_timexplain_2020}. 
However, both of these approaches rely on fixed window sizes. 
Thus they are incapable of modeling, e.g., semantically meaningful patterns in the time series, which can have variable lengths.
Additionally, they might miss important patterns if the predefined window size is smaller or longer than the pattern or if patterns lie between the fixed time segments.

Hence, we provide an extension of the LIME approach to identify superpixels-like patterns, i.e., semantically related data regions, in time series data. 
This paper presents a set of suitable segmentation algorithms and evaluates their suitability for providing explanations under various data characteristics.


\section{Post-hoc local explanations with LIME}
\begin{figure}
    \centering
    \includegraphics[width=0.95\linewidth]{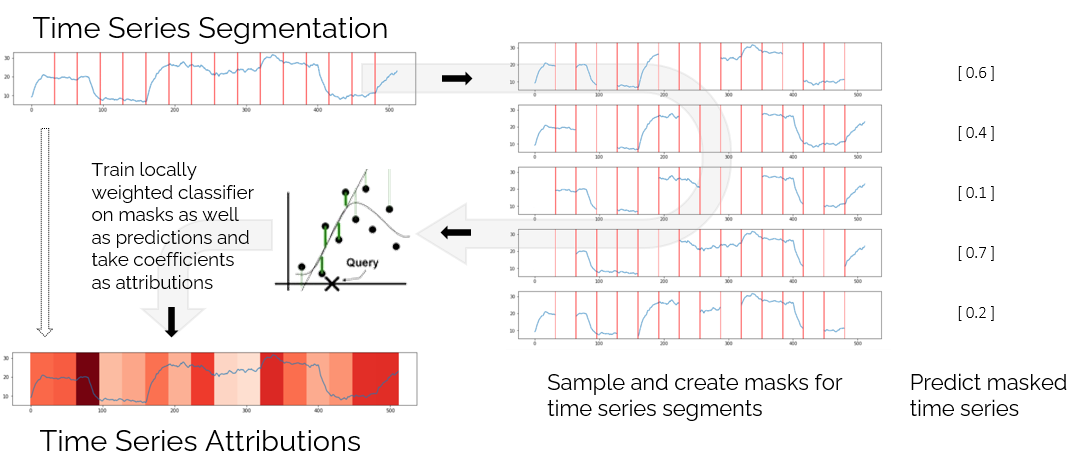}
    \vspace{-1em}
    \caption{The LIME approach applied on time series starting with the uniform segmentation on a time series sample. Next, doing the masking, perturbing, and predicting step of LIME to generate more local samples. Afterward, a linear interpretable model is trained on the masks and predictions using local weighting. At last, extracting the coefficients of the model leads to the wanted attributions for the initial sample.}
    \label{fig:limeonts}
    \vspace{-0.2em}
\end{figure}

Creating explanations for decisions of black-box models has various alternatives. 
One of these possibilities is the post-hoc approach LIME by Ribeiro et al.~\cite{ribeiro_why_2016}.
Local Interpretable Model-Agnostic Explanations, shortly LIME, uses an interpretable surrogate model to create explanations for black-box models.
In the first step, a chosen sample to explain and a model to be explained are given as input to the approach.
The sample is then segmented by a previously chosen segmentation algorithm, e.g., a superpixel segmentation for images~\cite{ribeiro_why_2016}.
LIME then creates masks for the sample deactivating segments or replacing them with non-informative values.
In many cases, this step is called perturbation and is something different than the perturbation mentioned later.
These newly generated (perturbed) samples are predicted with the input model to get new predictions.
LIME collects these predictions and trains a new interpretable classifier, often a linear model, on the masks with the predictions as the target.
In the case of a linear model, the coefficients are used to weigh the different input segments and to explain the model for the given sample.
Fig.~\ref{fig:limeonts} demonstrate the described approach on time series with a uniform segmentation.

LIME is generally applicable for any data type, but there are some challenges due to the necessity of segmentation.
Valuable segmentation makes sense to humans as it incorporates their domain knowledge.
For instance, superpixel segmentation identifies perceptual groups in images, which in most cases correspond to a human interpretable object.
As time series are generally hard to segment without domain knowledge, a general approach is rather difficult, even with domain knowledge not applicable.
A forecasting black box model often just uses a window as input to predict the target value in many cases.
Such a window is fixed beforehand and slides over the data, thus having no strict segmentation in itself.
Finding such segmentation is a significant challenge for time series as it needs to be generally applicable.


\section{Finding suitable segmentation mappings}
We propose \name, extending the LIME~\cite{ribeiro_why_2016} approach for time series with novel segmentation algorithms.
Our approach presents five segmentation techniques created for time series and three different replacement strategies.

\subsection{Using static windows}
\textit{Uniform segmentation} is the most basic method to segment a time series into windows. 
In this approach, we split the time series $ts = \{t_0, t_1, t_2, ..., t_n\}$ into equally and non-overlapping $m$-sized windows $ws = \{w_0, w_1, w_2, ..., w_{d}\}$ with $d = [n/m]$. 
If $n$ is not a multiple of $m$, the final windows may have more or less time points.
We expand the uniform segmentation to exponential windows, which ignores the size $m$ and has longer windows at the end. 
A time series $ts$ in \textit{exponential segmentation} is split into $d = log(n)$ windows and its length increases with {$[e^0], [e^1], [e^2], ..., [e^d]$}. 
To cover all the points of the time series, in the final window, we adjust its length by $n - \sum([e^0], [e^1], [e^2], ..., [e^{d-1}])$. 
A benefit of such segmentation is that we put more weight on the latest points with longer windows. 

\subsection{Using the Matrix Profile}
A matrix profile is a vector that stores the z-normalized Euclidean distance between any subsequence within a time series and its nearest neighbor~\cite{yeh_matrix_2016}.
Such a matrix profile can be used to identify motifs as well as outlier subsequences in large time series~\cite{yeh_matrix_2016}.
We introduce the \textit{slope} and \textit{bin} segmentation based on the matrix profile on time series to incorporate local trends and patterns.

The \textit{slope segmentation} has the parameters window size $m$ as input for the matrix profile and $k$ for the number of partitions for the segmentation.
The basic idea behind this segmentation approach consists in the opportunity to find patterns in the time series using the matrix profile.
By further focusing on the slope of the matrix profiles distances, we can identify drastic changes in the nearest neighbors to find not only possible patterns but also uncommon changes in the time series itself.
Such a uncommon changes can be used as plausible splits for the segmentation as the pattern are still included in the segments.
We calculate the matrix profile $mp = \{d_1, d_2, ..., d_j\}$ with our previously adjusted window size $m$ so that $j = n - m + 1$ to find interesting distances, e.g., to identify motifs. 
Afterward, we either calculate the gradient on the resulting matrix profile $\nabla mp$ and take the absolute value $|\nabla mp|$ to identify peaks as steep slopes.
Or, depending on the configuration, we sort the resulting matrix profile vector ascending and compute the slope to identify jumps in distances to find significant changes in the time series.
We sort the resulting vector in both cases and take the $k$-largest values to find segment borders.
The time series indices of these $k$ values segment our time series and describe drastic changes in the time series.

We further present \textit{bin segmentation} based on the matrix profile with the same parameters $m$ and $k$ as above.
Again the idea behind this approach enables finding patterns in the time series by not using the gradient to find drastic changes in the nearest neighbor but using bins to combine similar distances in the matrix profile to segments.
We calculate and sort the matrix profile again. 
However, we further split the min-max range of the matrix profile into $k$-bins.
Afterward, we label the $k$-bins numerically so that lower numbers have a low and higher a high matrix profile.
We convert our matrix profile to the corresponding bin number and assign our base value to the ${max}$ or ${min}$ bin.
Next, we slide over the resulting profile with a window length $m$.
Due to the sliding window approach, a time point can be either in the segment ${seg}_i$ or ${seg}_j$.
For our \textit{bins-min segmentation}, we assign the time point $t_t$ to ${seg}_i$ if ${bin}_i$ is smaller than ${bin}_j$.
Our \textit{bins-max segmentation}, oppositely, uses the ${seg}_i$ if ${bin}_i$ is larger than ${bin}_j$.

\subsection{Using the SAX transformation}
\textit{SAX segmentation} introduces a segmentation based on horizontal binning of a time series with $k$ partitions as the parameter.
The basic idea behind this segmentation approach includes the changes in the range of the values by splitting the overall distribution of possible values into bins.
The SAX transformation~\cite{lin_symbolic_2003} converts a time series $ts$ into a sequence of symbols $sa = \{s_0, s_1, s_2, ..., s_n\}$ with $s_i\in \{a, b, c, ...\}$ based on a continuous binning of intervals in the vertical direction.
We incorporate a base number of bins $b = 3$ for the SAX algorithm and use repeating symbols as segments, e.g., $sa = \{a_1, a_2, a_3, b_1, b_2, a_1, c_1, c_2, c_3\}$ involves four segments leading to $\{a_1, a_2, a_3\}, \{b_1, b_2\}, \{a_1\}, \{c_1, c_2, c_3\}$.
At each iteration, the amount of bins is increased $b += 1$ to finally achieve a previously selected $k$ partitions as more bins generally convert to more partitions.
For some cases, the exact partition size is not possible, and we allow a difference of ten percent to the selected partition size to mitigate such edge cases.

\subsection{Comparing the segmentation algorithms}
\begin{figure}[ht]
    \centering
    \includegraphics[width=0.9\linewidth]{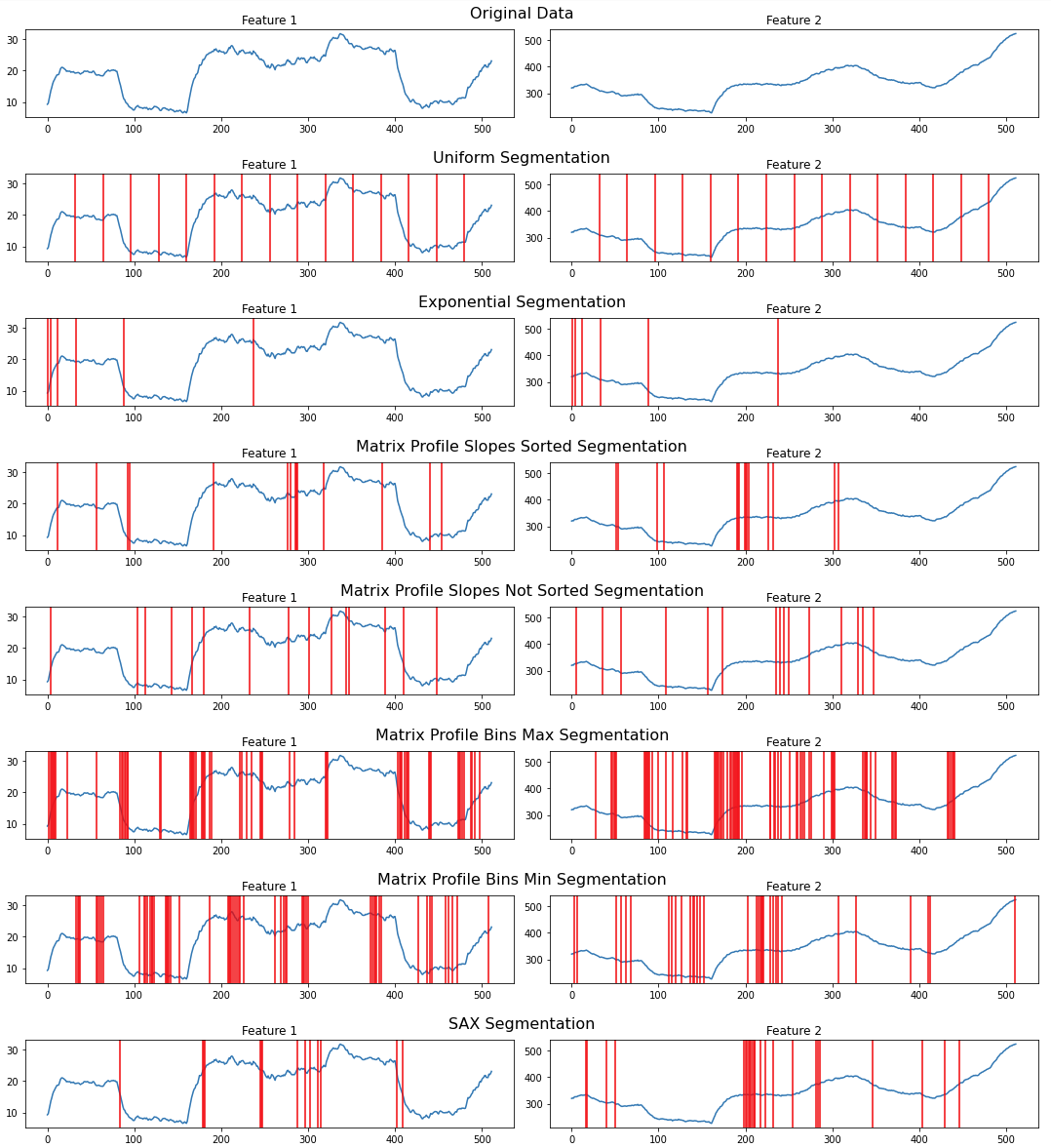}
    \vspace{-0.7em}
    \caption{Comparison of the different segmentation variants. Red stripes show segment splits. Some segmentation algorithms proposed end up with more as well as very short segments than the uniform segmentation with default parameters.}
    \label{fig:segmentation_comparison}
    \vspace{-0.2em}
\end{figure}

Existing and proposed segmentation algorithms lead to different segments representing potentially suitable techniques for various data sets.
Fig.~\ref{fig:segmentation_comparison} presents these algorithms on two differently scaled time series features.
Especially, comparing the uniform segmentation with the others demonstrates the advantages of the other approaches.
Depending on the algorithm, different segments are visible and present some more focused parts of the time series samples.
Choosing from a broader range of techniques can lead to improved explanations for humans.

\section{Evaluating \name on time series forecasting}
The evaluation of our proposed segmentation and perturbation approaches is based on the perturbation analysis for fidelity by Schlegel et al.~\cite{schlegel_towards_2019,schlegel_empirical_2020} adapted to forecasting tasks using the mean squared error.
As datasets for our evaluation, we use the Beijing Air Quality 2.5, Beijing Multi-Site Air Quality, and the Metro Interstate Traffic data to show the results on divers multivariate time series.
For the air quality datasets, we use a fixed input size of 24.
The metro traffic forecasting has an input length of 72.
We use three different basic implementations of black-box models: a basic one-dimensional convolutional neural network, a deep neural network, and a recurrent neural network (LSTMs~\cite{hochreiter_long_1997}).


The perturbation analysis by Schlegel et al.~\cite{schlegel_empirical_2020} consists of three steps: explanation generation, data perturbation based on explanations, and perturbation evaluation.
At first, a selected dataset, e.g., the test data, is evaluated with a quality metric (e.g., accuracy), and explanations are generated for every sample.
Next, every sample of the selected dataset is perturbed such that time points with high relevances for the explanation are replaced with non-information holding values.
As non-information holding values for time series are challenging to find, we focus on the proposed ones (zero, inverse, mean) by Schlegel et al.~\cite{schlegel_empirical_2020}.
Often the high relevance attributions are identified by using a threshold.
Lastly, the perturbed data gets evaluated, and the quality metric change is calculated.
The assumption is that a value change of the predicted data at highly relevant input positions decreases the quality metric performance of the model as the data loses valuable information.
Such an assumption then leads to the conclusion that a working XAI technique decreases the performance more than a random change.

{\renewcommand{\arraystretch}{1.2}%
\begin{table*}[]
\resizebox{\columnwidth}{!}{%
\centering\begin{tabular}{l|l|ccc|l|ccc|l|ccc}
        Zero    & \multirow{7}{*}{\rotatebox[origin=c]{90}{Beijing Air Quality 2.5}} & CNN & DNN & RNN & \multirow{7}{*}{\rotatebox[origin=c]{90}{Beijing AQ Multi Site}} & CNN & DNN & RNN & \multirow{7}{*}{\rotatebox[origin=c]{90}{Metro Interstate Traffic}} & CNN & DNN & RNN \\ \cline{1-1} \cline{3-5} \cline{7-9} \cline{11-13} 
Uniform     &                  &   \textbf{2.31}  &   \textbf{4.24}  &   2.32  &                         &   1.50  &  9.00   &  7.67   &                    &   2.43  &   0.22   &  6.55  \\
Exponential &                  &     0.56 & 1.12 & 1.41     &                         &     0.62 & 0.16 & \textbf{11.52}     &                    &     0.55 & 0.01 & 0.62    \\
Slopes      &                  &     1.31 & 2.11 & 1.95     &                         &    1.3 & 6.76 & 3.97     &                    &     \textbf{3.39} & 0.18 & \textbf{9.29}     \\
Bins Min    &                  &     0.35 & 3.43 & \textbf{3.6}     &                         &     0.41 & \textbf{10.46} & 5.71     &                    &     1.25 & 0.4 & 7.38     \\
Bins Max    &                  &     1.69 & 1.22 & 2.38     &                         &     \textbf{1.52} & 1.68 & 2.67     &                    &     1.44 & 0.44 & 2.68     \\
SAX         &                  &     1.24 & 2.58 & 2.23     &                         &     1.10 & 8.00 & 4.15    &                    &     1.55 & \textbf{1.16} & 7.34     \\
\end{tabular}
}
\vspace{0.5em}
\caption{
Evaluation results of the perturbation analysis for every segmentation technique for three datasets and three models. We calculate the perturbation analysis results based on the percentage change to the original prediction and the randomized change. A larger value shows a better explanation. 
}
\vspace{-0.8em}
\label{tab:eval}
\end{table*}
}

We extend the assumptions to calculate a score for improved comparability of the results by focusing on the percentage increase in relation to a random change of the time series.
Schlegel et al.~\cite{schlegel_empirical_2020} propose to take the 90th percentile value of the attribution values of the sample as a threshold.
However, we have to scale our \name values because we observed that depending on the segment count, the distribution of the attribution changes.
Such a distribution change leads to either more or less highly relevant time points for the perturbation as, e.g., there are more attribution values above the threshold value.
Thus, we take the initial prediction scores ${orig}$, the perturbed prediction scores ${pert}$, and the random position change prediction score ${rand}$ and calculate the increase of the perturbed: ${pert}_c = \frac{{pert} - {orig}}{{orig}}$ and random: ${rand}_c = \frac{{rand} - {orig}}{{orig}}$.
We set these in relation to get our final score: ${score} = \frac{|{pert}_c|}{|{rand}_c|}$.
A score below one depicts a worse performance than random guessing.
Scores larger than one illustrate plausible explanations better than guessing.
Through this scaling, the segmentation algorithms can be compared.
Larger results demonstrate better segmentation. 
Table~\ref{tab:eval} presents such a perturbation analysis on fidelity with our proposed segmentation approaches.

Our preliminary results for a zero perturbation, see Table~\ref{tab:eval}, show that \textit{uniform} is working well for short time series windows (Beijing Air Quality with 24) while \textit{slopes} generate better performances on long windows (Metro Interstate Traffic with 72).
However, also our proposed \textit{bins-min}, \textit{bins-max} and \textit{SAX} illustrate promising results for short windows and can be further tuned by adding more parameters.
Also, by further adding a minimum length for segments, these algorithms can be improved.
The DNN for the Metro Interstate Traffic dataset is interesting as non of the proposed segmentation strategies seem to work. 
However, such an effect can be caused as the model's performance is way worse than the other two models.
In general, the \textit{uniform} segmentation works well as a starting point, but exchanging it with our proposed algorithms enables more diverse and improved attributions. 


\section{Conclusion}
We present \name, a local interpretable model-agnostic explanation extraction technique for time series.
For \name, we extend the LIME approach with novel time series segmentation techniques and replacement methods to enforce a better non-informed values exchange.
Thus, we contribute five novel time series segmentation algorithms and the \name framework for time series forecasting.
We show on three forecasting datasets that \name performs better than randomly perturbing data and thus reveals relevant input values for the prediction of a model.
Further, we demonstrate that our proposed segmentation algorithms lead to improved attributions in most cases.
As future work, we want to compare the performance of \name against other XAI techniques applied to time series in the framework of Schlegel et al.~\cite{schlegel_empirical_2020}.
We also want to identify shapelets to generate segments with more in-depth domain knowledge and to investigate into similar attribution techniques like SHAP~\cite{lundberg_unified_2017}.

\subsection*{Acknowledgements}
This work has received funding from the European Union’s Horizon 2020 research and innovation programme under grant agreement No. 826494.

%
%
%
%
\bibliographystyle{unsrt}
\bibliography{main_paper}
\end{document}